\documentclass[a4paper,fleqn]{cas-sc}

\usepackage[numbers]{natbib}
\usepackage{csquotes}
\usepackage{amsmath}
\usepackage{mathtools}
\usepackage{float}

\def\tsc#1{\csdef{#1}{\textsc{\lowercase{#1}}\xspace}}
\tsc{WGM}
\tsc{QE}
\tsc{EP}
\tsc{PMS}
\tsc{BEC}
\tsc{DE}
\def\etal{\emph{et al.}}
\def\eg{\emph{e.g.}}
\begin{document}
\let\WriteBookmarks\relax
\def\floatpagepagefraction{1}
\def\textpagefraction{.001}
\shorttitle{Reviving Iterative Training with Mask Guidance for Interactive Segmentation}
\shortauthors{K.Sofiiuk, I.Petrov, A.Konushin}
%\begin{frontmatter}

\title [mode = title]{Reviving Iterative Training with Mask Guidance for Interactive Segmentation}

\author{Konstantin Sofiiuk}
\ead{k.sofiiuk@samsung.com}
\address{Visual Understanding lab., AI Center Moscow, Samsung Electronics Co., Lesnaya 5C, Moscow, Russia}

\author{Ilia A. Petrov}[orcid=0000-0002-8900-1071]
\cormark[1]
\ead{ilia.petrov@samsung.com}

\author{Anton Konushin}[orcid=0000-0002-6152-0021]
\cormark[2]
\ead{a.konushin@samsung.com}

\cortext[cor1]{Corresponding author}
\cortext[cor2]{Principal corresponding author}

%%%%%%%%% ---------------------------- ABSTRACT ----------------------------
\begin{abstract}
Recent works on click-based interactive segmentation have demonstrated state-of-the-art results by using various inference-time optimization schemes. These methods are considerably more computationally expensive compared to feedforward approaches, as they require performing backward passes through a network during inference and are hard to deploy on mobile frameworks that usually support only forward passes. In this paper, we extensively evaluate various design choices for interactive segmentation and discover that new state-of-the-art results can be obtained without any additional optimization schemes. Thus, we propose a simple feedforward model for click-based interactive segmentation that employs the segmentation masks from previous steps. It allows not only to segment an entirely new object, but also to start with an external mask and correct it. When analyzing the performance of models trained on different datasets, we observe that the choice of a training dataset greatly impacts the quality of interactive segmentation. We find that the models trained on a combination of COCO and LVIS with diverse and high-quality annotations show performance superior to all existing models. The code and trained models are available at \url{https://github.com/saic-vul/ritm_interactive_segmentation}.
\end{abstract}

%%%%%%%%% ------------------------------ KEYWORDS ------------------------------
\begin{keywords}
interactive segmentation \sep segmentation \sep mask refinement
\end{keywords}

\maketitle

%%%%%%%%% ---------------------------- INTRODUCTION ----------------------------
\section{Introduction}

Interactive segmentation algorithms allow users to explicitly control the predictions using interactive input at several iterations, in contrast to common semantic and instance segmentation algorithms that can only input an image and output a segmentation mask in one pass. Such interaction makes it possible to select an object of interest and correct prediction errors. Another important feature of this group of algorithms is the capability to segment objects of previously unseen classes.

User input can be formalized via various representations: scribbles, clicks, extreme points, etc. Click-based interactive segmentation is one of the most well-studied topics among other deep learning-based interactive segmentation approaches, as it has well-established protocols of training and evaluation \cite{xu2016deep, jang2019interactive, kontogianni2020continuous, sofiiuk2020f}. It also employs quite an intuitive and simple way to specify the desired object. Scribble-based methods often use heuristic and complicated procedures for simulating user input, which makes it hard to fairly evaluate them \cite{bai2014error, lin2016scribblesup, freedman2005interactive}. Approaches based on extreme points are not intuitive for users and are not flexible enough due to a limited number of user interactions \cite{maninis2018deep}. In our work, we focus on click-based interactive segmentation.

\begin{figure}
    \centering
    \includegraphics[width=0.8\linewidth]{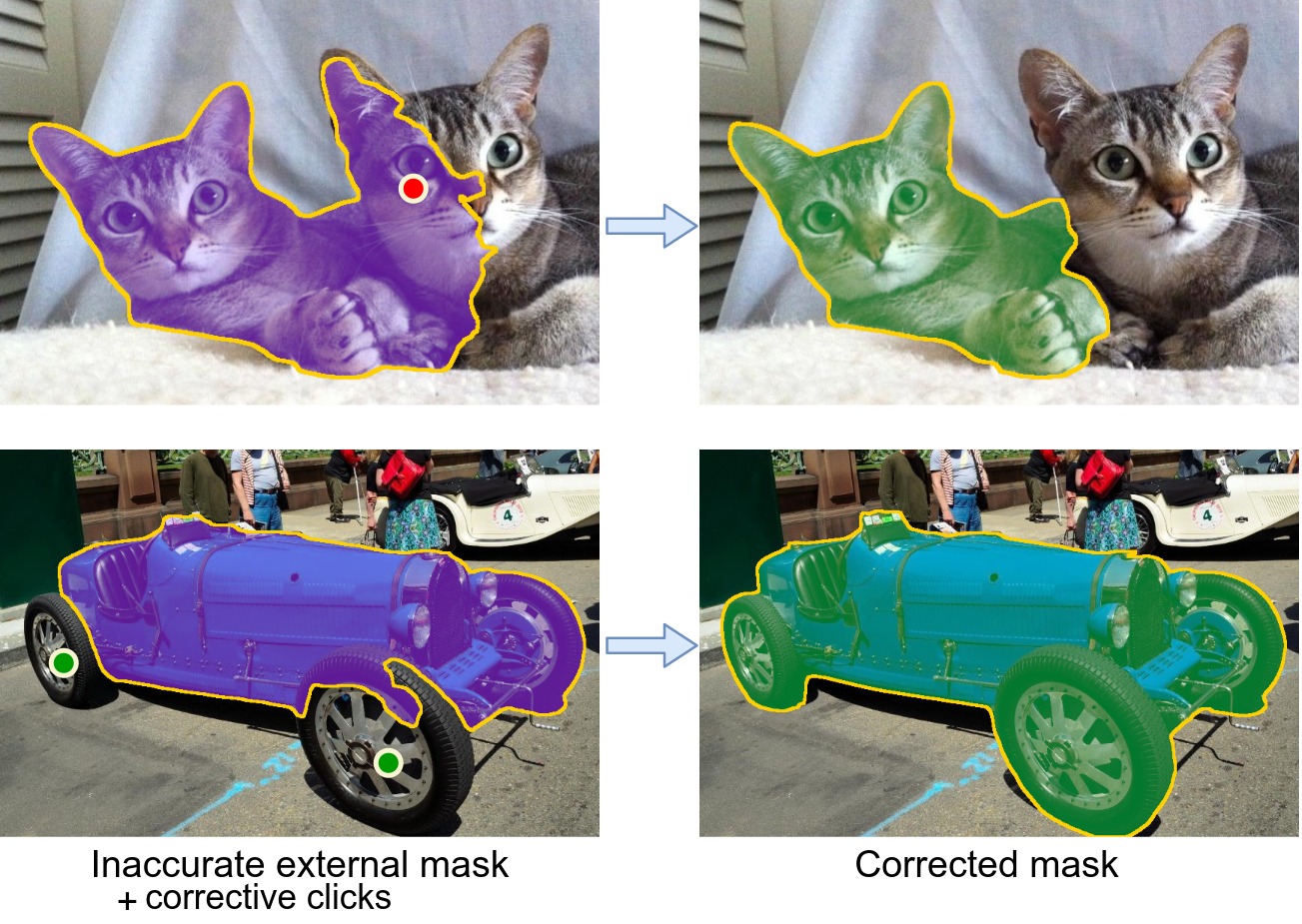}
    \caption{Besides segmenting new objects, proposed method allows to correct external masks, \eg produced by other instance or semantic segmentation models. A user can fix false negative and false positive regions with positive (green) and negative (red) clicks, respectively.}
    \label{fig:external_mask_correction}
\end{figure}

In general, the development of deep learning-based algorithms for semantic and instance segmentation requires a huge amount of annotated data.  Annotating data with segmentation masks is very time-consuming and therefore expensive. Interactive segmentation can significantly simplify and speed up this process \cite{acuna2018efficient, benenson2019large, agustsson2019interactive}, which is one of the most important applications of interactive segmentation. It can also be used in photo editing, allowing users to select objects easily, which is especially important for smartphone applications where user input is often limited by finger or stylus activities.

Recent works on click-based interactive segmentation propose complicated inference-time optimization procedures to improve the quality of interactive segmentation even further \cite{jang2019interactive, kontogianni2020continuous, sofiiuk2020f}. The deployment of these models is substantially limited, for instance, on mobile devices, as it requires the implementation of backward passes with gradients that are not provided by popular frameworks. Surprisingly, we find that properly chosen baselines of click-based interactive segmentation models without any without any explicit refinement techniques and other specific modifications can be trained using the standard random points sampling procedure and show state-of-the-art performance. We observe that modifications of poorly tuned baselines can provide false performance improvements, while strong baselines with the same modifications do not show any improvement at all. At the same time, datasets with coarse masks or containing a small amount of images may become a bottleneck for the performance of interactive segmentation models. Our experiments show that usage of diverse large datasets with fine masks for training plays a crucial role for the performance of discussed models. Therefore, we choose a strong baseline model for implementing and evaluating new modifications. We train models on a combination of LVIS and COCO datasets \cite{lin2014microsoft, gupta2019lvis}, that, to the best of our knowledge, are the most suitable for training interactive segmentation models.

We propose an extension of click-based interactive segmentation that allows to modify existing instance segmentation masks interactively. We revive an iterative training procedure, and make a network aware of the mask from a previous step \cite{mahadevan2018iteratively}. We show that such awareness improves the models' stability, i.e. allows to avoid accuracy dropping when adding new clicks. We propose a new training dataset obtained by combining the LVIS and COCO datasets.

%%%%%%%%% ---------------------------- RELATED WORK ----------------------------
\section{Related Work}
\label{sec:related}

\textbf{Interactive segmentation methods.} Interactive segmentation is a longstanding problem in computer vision. Early methods \cite{boykov2001interactive, grady2006random, gulshan2010geodesic, rother2004grabcut} tackle the problem using optimization-based approaches minimizing a specifically constructed cost function defined on a graph over image pixels. GrabCut proposed in \cite{rother2004grabcut} is a classic approach based on iterative energy minimization of a cost function, that is modeled using a Gaussian mixture. Xu~\etal~\cite{xu2016deep} first propose a CNN-based model for interactive segmentation and introduced a clicks simulation strategy for training that is adopted in some further works. Later, \cite{li2018interactive, liew2017regional, liew2019multiseg, lin2020interactive} propose various CNN-based methods for interactive segmentation that aimed at the refinement of predictions by increasing the diversity of predicted masks, and using the attention mechanism. The novel subgroup of methods emerges with an introduction of the Backpropagating Refinement Scheme (BRS) in \cite{jang2019interactive}. The authors propose an optimization procedure that minimizes a discrepancy between the predicted mask and the map of input clicks after each click with respect to the input distance maps. This refinement technique improves the segmentation quality at the cost of increasing the runtime. Sofiiuk~\etal~\cite{sofiiuk2020f} address this issue by proposing f-BRS, a lightweight version of BRS that uses a similar optimization procedure, though with respect to internal parameters introduced to the higher levels of a network, reducing runtime on an order of magnitude. Kontogianni~\etal~\cite{kontogianni2020continuous} introduce another method of test time refinement targeting the optimization process on network parameters.

\textbf{Different types of interactive feedback.} While clicks are the mainstream form of input in interactive segmentation, a lot of works explore other variations of interactive feedback from the user. For example, one of the simplest types of interactions is a bounding box, that is used in \cite{cheng2015densecut, rother2004grabcut, wu2014milcut}. The main drawbacks of these approaches are lack of specific object reference inside the selected area and lack of interface for correction of the predicted mask. The limitations of bounding box-based interaction are addressed in \cite{zhang2020interactive}. The authors propose to combine clicks with bounding boxes to provide more specific object guidance and allow corrections of the predicted mask. DEXTR \cite{maninis2018deep} uses extreme points of the target object, i.e. left-most, right-most, top, and bottom pixels as an input. On the one hand, such an interaction is compact and is limited by 4 clicks. On the other hand, placing extreme points in the right locations is harder than making an arbitrary click on an object and there is no support for corrections, similar to interaction via bounding boxes. Scribbles are used in many early works \cite{bai2014error, boykov2001interactive, grady2006random, gulshan2010geodesic, kim2008generative}. This type of feedback provides richer prior information for the algorithm, compared to the others. In contrast, putting a stroke requires more effort from the user, compared to simpler forms of interactions. Another drawback that refrains CNN-based methods from the wide adoption of scribbles is that realistic strokes simulation for training the neural networks is a rather hard task. Therefore, the vast majority of scribble-based methods employ graphical models or similar training-free techniques. Apart from the aforementioned forms of interaction, some works use a combination of traditional input representation with other modalities, e.g. PhraseClick \cite{ding2020phraseclick} combines clicks with text input to better infer the attributes of the target object, thus requiring fewer clicks.

\textbf{Segmentation mask refinement.} The majority of segmentation mask refinement techniques are targeted to refine the boundaries of a mask by making local corrections without changing the mask globally, i.e. excluding or including large parts of an object. These methods can either be a part of the network architecture \cite{pinheiro2016learning, yuan2020segfix} or serve as a post-processing step \cite{chen2017deeplab, zhou2020deepstrip}. Local and boundary refinement tasks are often addressed without any additional input from the user, as all the required features are automatically extracted from the input image and mask. Global refinement, on the contrary, may require additional feedback from the user to make improvements to the mask. We propose to combine interactive segmentation and mask refinement allowing to refine external segmentation mask with user clicks as guidance.

%%%%%%%%% ------------------------ PROPOSED METHOD ----------------------------
\section{Proposed Method}
\label{sec:proposed}

\subsection{Revising Network Architecture}
\label{sec:revising_network_and_brs}

\begin{figure}
    \centering
    \includegraphics[width=0.6\linewidth]{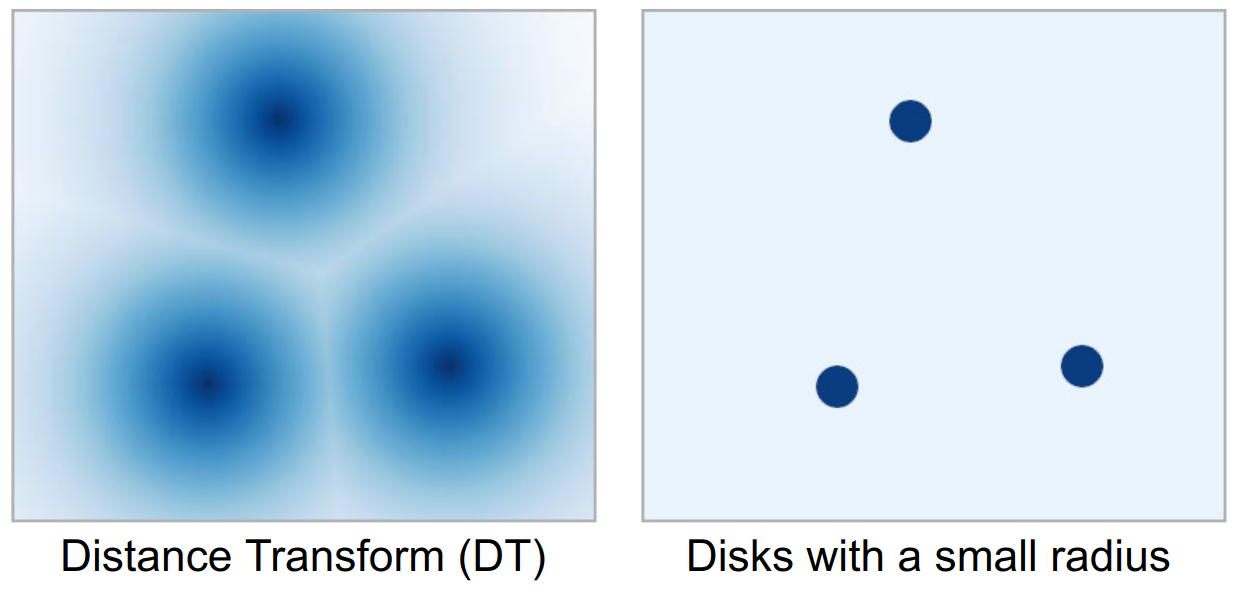}
    \caption{Visualization of two different approaches for encoding user clicks.}
    \label{fig:clicks_encodings}
\end{figure}

The task of interactive segmentation is very similar to instance or semantic segmentation in terms of the network architecture. In these tasks, networks take high-resolution input and produce high-resolution segmentation masks aligned with the input. The key difference is in the user input: its main aspects are the encoding and processing of the encoded input inside the network. Therefore, there is no need to reinvent general segmentation architectures. Instead, it is reasonable to rely on time-tested state-of-the-art segmentation networks and focus on interactive-specific parts.

We consider the DeepLabV3+~\cite{chen2018encoder} and HRNet+OCR~\cite{WangSCJDZLMTWLX19, yuan2019object} semantic segmentation architectures as a backbone for our interactive segmentation model. While DeepLabV3+ is a well-studied segmentation architecture that proved efficient in many segmentation-related tasks, HRNet is a relatively new promising one that was specially designed to produce high-resolution output. According to our experiments, HRNet is a more preferable architecture for this task. We provide the ablation study for the backbone architecture in Section~\ref{sec:ablation} and show the results in Tables~\ref{tab:ablation_architecture}, \ref{tab:ablation_datasets}~and~\ref{tab:arch_comparison}. 

\begin{figure}[t]
    \centering
    \includegraphics[width=0.6\linewidth]{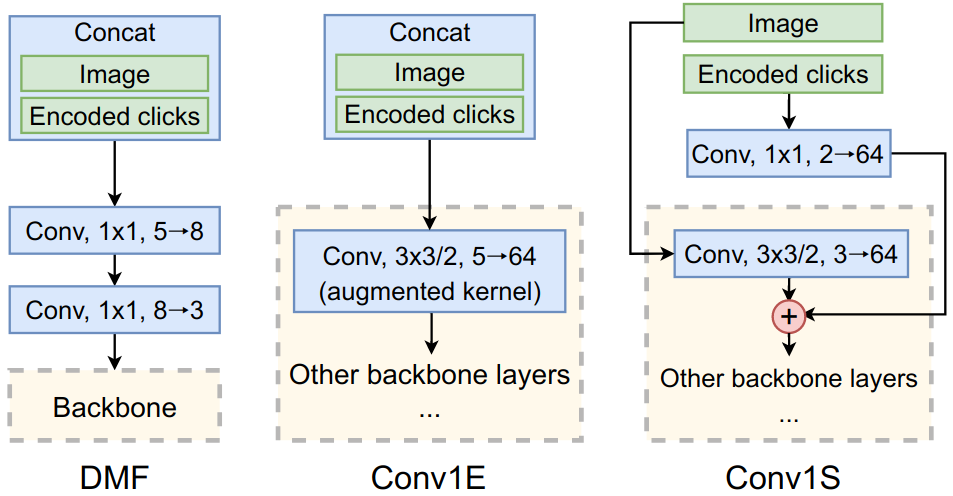}
    \caption{Different architecture choices of feeding encoded clicks to a backbone, described in Section~\ref{sec:revising_network_and_brs}.}
    \label{fig:input_schemes}
\end{figure}

\textbf{Clicks encoding.} There are positive and negative clicks in click-based interactive segmentation. These clicks are represented by their coordinates in an image. In order to feed them to a convolutional network, we should encode them in a spatial form first. Encoding the clicks via a distance transform from Xu~\etal~\cite{xu2016deep} is the most common approach to clicks encoding. However, they can also be represented by gaussians or disks with a fixed radius. Benenson~\etal~\cite{benenson2019large} perform a detailed ablation study on clicks encoding and find that disks with a small radius surpass the other encodings in terms of model performance.

We conduct our ablation study to compare the distance transform encoding with the disks encoding and find that the latter outperforms the former. We assume that the following observation explains the superiority of disks. The changes in disk encoding caused by adding new points or moving existing ones are always local and only slightly affect the encoding map. At the same time, a distance transform map can change drastically when a new point is added, especially if there are only a few points. In turn, such sudden considerable changes might confuse a network. We show a visualization of both the distance transform and disk encodings in Figure~\ref{fig:clicks_encodings}.

\textbf{Feeding encoded clicks to a backbone.} Semantic segmentation backbones are usually pre-trained on ImageNet~\cite{deng2009imagenet} and  can take only RGB images as an input. The most common way to handle additional input, e.g. encoded user clicks, is to augment the weights of the first convolutional layer of a pre-trained model to accept N-channel input instead of only an RGB image~\cite{xu2016deep, benenson2019large}. In our work, we denote this modification of network architecture by Conv1E. Sofiiuk~\etal~\cite{sofiiuk2020f} propose the Distance Maps Fusion (DMF) module which transforms an image concatenated with additional user input channels into 3-channel input.

We propose a new simple approach to solving this problem, which is described as follows. We introduce a convolutional block that outputs the tensor of exactly the same shape as the first convolutional block in the backbone does. This tensor is then summed element-wise with the output of the first backbone convolutional layer, which usually has 64 channels. We denote this modification of network architecture by Conv1S. While being similar to Conv1E, it allows to choose a different learning rate for new weights without affecting the weights of a pre-trained backbone. We show the schemes of these architectural modifications in Figure \ref{fig:input_schemes} and provide an ablation study in Section \ref{sec:ablation}.

% Table: comparison of architectures
\begin{table}
\footnotesize
\begin{center}
\caption{Ablation studies of the network architecture choices described in Section \ref{sec:revising_network_and_brs}. Each cell consists of two results "X/Y", where "X" and "Y" correspond to evaluation without and with \mbox{f-BRS-B\cite{sofiiuk2020f}}, respectively. "DT" stands for the distance transform clicks encoding. All models are trained on SBD.}
\begin{tabular}{l|c|c|c|c}
\hline
\multirow{2}{*}{Backbone} & 
\multirow{2}{*}{\begin{tabular}{@{}c@{}} Input \\ Scheme \end{tabular}} & 
\multirow{2}{*}{\begin{tabular}{@{}c@{}} Clicks \\ Encoding\end{tabular}} &
\multicolumn{2}{c}{NoC\textsubscript{20}@90} \\
\cline{4-5}
{} & {} & {} & Berkeley & DAVIS \\
\hline
\hline
\multirow{5}{*}{ResNet-34}        & DMF \cite{sofiiuk2020f}    & DT      & 5.50/4.32 & 8.45/8.34 \\
                                  & Conv1E & DT                & 4.79/4.43 & 7.56/7.60 \\
                                  & Conv1S & DT                & 4.98/4.16 & 7.41/7.28 \\
                                  & Conv1S & Disk3             & 4.52/4.04 & 7.27/7.18 \\
                                  & Conv1S & Disk5             & 4.09/3.89 & 6.92/7.22 \\
\hline
\multirow{5}{*}{HRNet-18}    & DMF \cite{sofiiuk2020f}   & DT      & 4.93/4.35 & 8.59/8.00 \\
                                 & Conv1E                & DT      & 4.41/3.95 & 7.50/7.43 \\
                                 & Conv1S                & DT      & 3.99/3.81 & 7.16/7.24 \\
                                 & Conv1S                & Disk3   & 3.63/3.47 & 7.14/7.04 \\ 
                                 & Conv1S                & Disk5   & 3.52/3.50 & 6.90/6.97 \\
\hline

\end{tabular}
\label{tab:ablation_architecture}
\end{center}
\end{table}

\subsection{Iterative Sampling Strategy}
\label{sec:iterative_sampling}

Most recent works on click-based interactive segmentation use the sampling strategy for simulating user clicks during training, where a set of positive and negative clicks is randomly generated without considering any relations between them \cite{sofiiuk2020f, lin2020interactive, benenson2019large, xu2016deep}. In practice, every new click is placed in the erroneous region of a prediction produced by a network using the set of previous clicks. This fact is completely ignored in the random sampling strategy. It also makes it impossible to integrate masks from previous interactions into a model, as we need to have ordered interactions and a sequence of corresponding predictions to successfully train such a model.

The iterative sampling strategy, which resembles the interaction with a real user, is employed to a certain extent in several works \cite{mahadevan2018iteratively, majumder2019content, kontogianni2020continuous}. As full iterative sampling is very computationally expensive, in these works random sampling is used for initialization and then a few clicks are added using the iterative sampling procedure.

We adopt the iterative sampling procedure proposed by Mahadevan~\etal~\cite{mahadevan2018iteratively} with the following changes. First, we sample each point not just from the center of a mislabelled region, but from the region obtained by applying morphological erosion operation to the mislabelled region, so that the eroded region has 4 times less area. We observe that choosing center points leads to overfitting to the NoC evaluation metric (see Section \ref{sec:experiments} for the details). In practice, however, these models demonstrate worse performance and unstable behavior when a user places a click near the borders of an object or a mislabelled region. Second, we do not save simulated clicks for dataset samples during training and simulate user clicks for each batch individually. For that reason, we limit the maximum number of sampling iterations to $N_{iters}$, and each batch can uniformly get from~0~to~$N_{iters}$ iterations.

We use a combination of the random and iterative sampling strategies for training our iterative models. First, we simulate user clicks with the random sampling strategy \cite{xu2016deep} just as we do for our non-iterative baselines. Then, we add from~0~to~$N_{iters}$ iteratively simulated clicks. We provide an ablation study of the number of iteratively sampled clicks in Section \ref{sec:ablation}.

\subsection{Incorporating Masks From Previous Steps}
\label{sec:masks_from_previous_steps}

In interactive segmentation, it seems natural to incorporate output segmentation masks from previous interactions as an input for the next correction, providing additional prior information that can help improve the quality of prediction. In case of incorporating the mask from a previous interaction, it is necessary to use the iterative sampling for simulating user interactions during training. Along with the iterative sampling, Mahadevan~\etal~\cite{mahadevan2018iteratively} propose passing an output mask from a previous iteration into the model as an optional extension of their method.

To train a model with the mask from a previous step, we use a combination of the random and iterative user interactions simulation described in Section~\ref{sec:iterative_sampling}. Our model takes this mask as the third channel together with two channels for positive and negative encoded clicks, respectively. For the first interaction, as well as for the batches with skipped iterative sampling, we feed an empty mask to our model.

\subsection{Normalized Focal Loss}
\label{sec:normalized_focal_loss}
Binary cross entropy (BCE) loss is one of the standard loss functions for training semantic segmentation algorithms. Besides, some state-of-the-art interactive segmentation methods \cite{jang2019interactive, liew2019multiseg, majumder2019content, maninis2018deep, zhang2020interactive} adopt it for training the network. The main drawback of BCE is that it treats all examples equally, slowing the training during later epochs as the gradient from almost correctly segmented areas propagates similarly to the gradient from erroneous regions. Focal loss was introduced in \cite{lin2017focal} to alleviate this problem. Let $\hat{M}$ denote the output of the network and $p_{i,j}$ denote the confidence of prediction at the point $(i,j)$. Then the FL is formulated as follows:

\begingroup

\setlength\abovedisplayskip{0pt}
\setlength\belowdisplayskip{6pt}
\begin{equation}
    FL(i,j)=-(1-p_{i,j})^{\gamma}\log p_{i,j}
\label{eq:FL}
\end{equation}

\endgroup

One can notice that the total weight $P(\hat{M})=\sum_{i,j}(1-p_{i,j})^\gamma$ decreases when the accuracy of the prediction increases. This means that the total gradient of the FL fades over time, slowing the training process. To mitigate this problem, \cite{sofiiuk2019adaptis} proposed normalized focal loss (NFL) that is formulated as follows:

\begingroup
\setlength\abovedisplayskip{0pt}
\setlength\belowdisplayskip{6pt}
\begin{equation}
    NFL(i,j,\hat{M})=-\frac{1}{P(\hat{M})}(1-p_{i,j})^{\gamma}\log p_{i,j}
\label{eq:NFL}
\end{equation}
\endgroup

The gradient of NFL does not fade over time due to normalization and remains equal to the total gradient of BCE. This allows for faster convergence and better accuracy compared to training with BCE. We choose NFL as a loss function for training the proposed methods and provide an ablation study with respect to loss function choice in Section~\ref{sec:ablation} (results are provided in Table~\ref{tab:ablation_loss}).

% Table: comparison of loss functions
\begin{table}
\small
\begin{center}
\caption{Evaluation of the non-iterative baseline method with the \mbox{HRNet-18+OCR} backbone trained on COCO+LVIS (proposed in Section \ref{sec:coco_lvis}) with different loss functions. We report NoC@90 on four datasets.}
\begin{tabular}{l|c|c|c|c}
\hline

\multirow{2}{*}{Method} &
\multicolumn{4}{c}{NoC\textsubscript{20}@90} \\
\cline{2-5}
{} & GrabCut & Berkeley & SBD & DAVIS \\
\hline
\hline
BCE      & 1.82 & 3.13 & 7.58 & 6.31 \\
Soft IoU & 2.02 & 3.03 & 7.94 & 6.45 \\
FL       & 1.80 & 3.28 & 7.56 & 6.40 \\
NFL      & \textbf{1.70} & \textbf{2.48} & \textbf{6.72} & \textbf{5.90} \\
\hline

\end{tabular}
\label{tab:ablation_loss}
\end{center}
\end{table}

%%%%%%%%% ------------------------ COCO+LVIS for IS ---------------------------
\section{Dataset for Interactive Segmentation}
\label{sec:dataset_for_is}

\subsection{Reviewing Existing Datasets}
\label{sec:existing_datasets}
The vast majority of recent CNN-based interactive segmentation methods \cite{jang2019interactive, li2018interactive, liew2017regional, liew2019multiseg, majumder2019content, sofiiuk2020f, xu2016deep} are trained either using Semantic Boundaries Dataset \cite{hariharan2011semantic}, or the Pascal~VOC dataset \cite{everingham2010pascal}, or the combination of these two datasets, as they share the same set of images. The augmented dataset contains a total of 10582 images with 25832 instance-level masks. The total number of classes in this dataset is twenty: 7 categories of transportation, 6 species of animals, 6 types of indoor objects, and a separate class for persons. These classes cover only general types of objects, implying limitations on the variety of predictable classes.
Recently, large-scale segmentation datasets OpenImages \cite{benenson2019large, kuznetsova2018open} and LVIS \cite{gupta2019lvis} were introduced. These datasets stimulated the development of state-of-the-art instance segmentation algorithms \cite{hu2020learning, jiawei2020balanced, tan2020equalization, wang2020solov2}, providing a large variety of labeled classes with a sufficient number of examples. LVIS contains around 1.2M instances on 100k images embracing more than a thousand object classes, while OpenImages has 2.6M instances on 944k images covering 350 object categories. We believe that having such a diverse dataset is one of the key components in training of the state-of-the-art interactive segmentation model.
Another important characteristic of a segmentation dataset is annotation quality. Gupta~\etal~\cite{gupta2019lvis} present a study of annotation quality in some of the instance segmentation benchmarks, comparing a subset of masks from each dataset to experts’ annotations. According to this study, the LVIS dataset has the highest annotation quality among reported datasets, presumably allowing to achieve higher prediction quality.

\subsection{Combination of COCO and LVIS}
\label{sec:coco_lvis}
Considering the observations from Section~\ref{sec:existing_datasets}, the LVIS dataset appears to be the best choice for training the models except for one drawback: it is long-tailed and therefore lacking general object categories, which can affect the accuracy and generalization of the trained model. As a solution to this problem, we propose to augment LVIS labels with masks from the COCO \cite{lin2014microsoft} dataset, as these two datasets share the same set of images. The COCO segmentation dataset contains a total of 1.2M instance masks on 118k training images with 80 object classes. These categories represent more common and general objects, complementing the long-tailed object classes from the LVIS dataset. We propose the following procedure to construct the combined COCO+LVIS dataset with diverse object classes from the most common to less frequent ones. All masks from both datasets are joined together except for those masks from COCO that have a corresponding mask from LVIS with an intersection over union (IoU) score between them larger than 80\%. In that case, we only keep the mask from LVIS, since it presumably has better overall and especially boundary quality.
As a result of the described procedure, we obtained a dataset with 104k images and 1.6M instance-level masks. We argue that the further development of interactive segmentation algorithms relies heavily on the training data. Thus, we provide the comparative study of the training datasets in Section \ref{sec:ablation}, results are provided in Table~\ref{tab:ablation_datasets}. Based on our experiments and the aforementioned observations, we conclude that the proposed COCO+LVIS dataset is the best choice for training interactive segmentation methods.

% Table: comparison of training datasets
\begin{table}
\footnotesize
\begin{center}
\caption{Evaluation results for the non-iterative baseline method with the \mbox{ResNet-34} and \mbox{HRNet-18+OCR} backbones trained on various datasets. NoC@90 is reported.}
\begin{tabular}{l|l|c|c|c|c}
\hline
\multirow{2}{*}{Train Dataset} &
\multirow{2}{*}{Backbone} & 
\multicolumn{4}{c}{NoC\textsubscript{20}@90} \\
\cline{3-6}
{} & {} & GrabCut & Berkeley & SBD & DAVIS \\
\hline
\hline
\multirow{2}{*}{ADE20k}        & ResNet-34    & 2.70 & 5.09 & 8.27 & 8.56 \\
                               & HRNet-18     & 2.68 & 4.78 & 8.02 & 8.32 \\
\hline
\multirow{2}{*}{OpenImages}    & ResNet-34    & 2.20 & 4.68 & 8.16 & 7.47 \\
                               & HRNet-18     & 2.02 & 4.47 & 7.95 & 8.08 \\
\hline
\multirow{2}{*}{SBD}           & ResNet-34    & 2.94 & 4.73 & 6.94 & 7.56 \\
                               & HRNet-18     & 2.41 & 3.95 & 6.66 & 7.17 \\
\hline
\multirow{2}{*}{Pascal~VOC+SBD} & ResNet-34   & 2.77 & 4.53 & 6.92 & 6.48 \\
                                & HRNet-18    & 2.25 & 3.63 & 6.63 & 6.16 \\
\hline
\multirow{2}{*}{LVIS}          & ResNet-34    & 2.59 & 3.61 & 7.99 & 6.98 \\
                               & HRNet-18     & 2.44 & 3.13 & 8.14 & 7.18 \\
\hline
\multirow{2}{*}{COCO}          & ResNet-34    & 1.80 & 3.34 & 6.29 & 6.11 \\
                               & HRNet-18     & 1.77 & 2.90 & 6.32 & 5.85 \\
\hline
\multirow{2}{*}{COCO+LVIS}     & ResNet-34    & 1.74 & 2.91 & 6.53 & 6.01 \\
                               & HRNet-18     & 1.70 & 2.48 & 6.86 & 6.00 \\
\hline

\end{tabular}
\label{tab:ablation_datasets}
\end{center}
\end{table}

%%%%%%%%% ---------------------------- EXPERIMENTS ----------------------------
\section{Experiments}
\label{sec:experiments}

We perform an extensive evaluation of the proposed approach by conducting ablation studies for all sufficient parts of the method, exploring the convergence properties with an increasing number of clicks and comparing our method with current state-of-the-art works.

\textbf{Datasets.}
We evaluate the performance of our method on five common benchmarks for interactive segmentation with instance-level annotations. The GrabCut \cite{rother2004grabcut} dataset contains 50 images with a single object mask for each image. We adopt the test subset of Berkeley \cite{martin2001database} introduced in \cite{mcguinness2010comparative}, which consists of 100 masks for 96 images. The DAVIS \cite{perazzi2016benchmark} dataset was introduced for the evaluation of video segmentation datasets. We use the subset of 345 randomly sampled frames of video sequences that was introduced in \cite{jang2019interactive} for evaluation. We follow the common protocol and combine all instance-level masks for one image into one segmentation mask. We also use the validation part of the Pascal VOC \cite{everingham2010pascal} dataset that consists of 1449 images with 3417 instances. Each instance mask is used separately during evaluation. The Semantic Boundaries Dataset (SBD) \cite{hariharan2011semantic} contains 6671 instance-level masks for 2820 images. This dataset has been used for evaluating the interactive segmentation algorithms since \cite{xu2016deep}. To test the algorithm, we use each of instance masks separately and do not combine them into one segmentation mask for one image.

\textbf{Evaluation metric.}
We perform the evaluation using the standard Number of Clicks (NoC) measure, reporting the number of clicks required to achieve the predefined Intersection over Union (IoU) threshold between predicted and ground truth masks. We denote NoC with IoU threshold set to 85\% and 90\% as NoC@85 and NoC@90, respectively. To generate clicks during the evaluation procedure, we follow the strategy used in \cite{li2018interactive, xu2016deep}. The next click is placed at the center of the region with any type of prediction error (false positive or false negative) with the largest area among other erroneous regions. The region center is defined as the point farthest from the boundaries of the corresponding region.

% Table: comparison of models capacity and complexity
\begin{table}
\small
\begin{center}
\caption{Comparison of different models in terms of the number of FLOPs and parameters. All models take an image with a~resolution of $400\times400$ to compute the number of FLOPs. We use exactly the same architecture of the DeepLab models as it was proposed in \cite{sofiiuk2020f}.}
\begin{tabular}{l|cc|cc}
\hline

Model & \#Params & Ratio-to-HRNet18 & \#FLOPs & Ratio-to-HRNet18 \\
\hline
\hline
HRNet18s+OCR           & 4.22M  & 0.4x & 17.84G  & 0.6x   \\
HRNet18+OCR            & 10.03M & 1.0x & 30.80G   & 1.0x   \\
HRNet32+OCR            & 30.95M & 3.1x & 82.84G  & 2.7x   \\
DeepLab-ResNet-34      & 19.17M & 1.9x & 122.28G & 4.0x   \\
DeepLab-ResNet-50      & 31.40M & 3.1x & 170.13G & 5.5x   \\
\hline

\end{tabular}
\label{tab:arch_comparison}
\end{center}
\end{table}

\textbf{Segmentation backbones.}
We consider two different segmentation architectures: DeepLab-V3+ \cite{chen2018encoder} with ResNet and HRNet+OCR \cite{WangSCJDZLMTWLX19, yuan2019object}. To implement DeepLab-V3+ for interactive segmentation we follow \cite{sofiiuk2020f}. We use the implementation of HRNet and models pre-trained on ImageNet presented in the official repository\footnote{\url{https://github.com/HRNet/HRNet-Image-Classification}}. There are several versions of HRNet which differ in terms of their capacity: HRNet-W18-C, HRNet-W30-C, HRNet-W32-C, etc. In our work we employ HRNet-W18-C-Small-v2, HRNet-W18-C, HRNet-W32-C, referring to them as HRNet-18s, HRNet-18, HRNet-32 for the sake of brevity. The numbers of parameters and FLOPs of our models is shown in Table \ref{tab:arch_comparison}. All the networks are initialized by the weights pre-trained on ImageNet weights.

We use the OCR module proposed in \cite{yuan2019object} for all HRNets. The original OCR module always produces 512 output channels regardless of the size of a backbone. We proportionally reduce the width of all the OCR layers, so that it produces 48, 64 and 128 channels for HRNet-18s, HRNet-18 and HRNet-32, respectively.

\textbf{Implementation details.}
The training task is binary segmentation with normalized focal loss, described in Section \ref{sec:normalized_focal_loss}, as an objective. We use image crops with a size of $320\times480$ and randomly resize images with a scale factor from 0.75 to 1.40 before cropping. We use horizontal flip and random jittering of brightness, contrast, and RGB values as augmentations during training. We adopt test time augmentations from f-BRS method \cite{sofiiuk2020f}, using Zoom-In technique and averaging predictions from original and horizontally flipped image during evaluation.

In all our experiments, we use Adam optimizer with $\beta_1 = 0.9, \beta_2 = 0.999$ and train the models for 55 epochs on the proposed COCO+LVIS dataset (49 epochs with learning rate $5 \times 10^{-4}$, then learning rate is decreased by 10 times on 50th and 53rd epochs). The learning rate for backbone networks is set 10 times lower than the learning rate for the rest of the network. We denote one pass through each image of the dataset as an epoch. The batch size is set to 32. We train the networks based on HRNet-18s, HRNet-18 and HRNet-32 on 1, 2 and 4 Tesla P40, respectively. To train the model based on ResNet-34, we use 1 Tesla P40.

We implement the proposed method using the PyTorch \cite{paszke2019pytorch} framework.

\subsection{Convergence Analysis}
\label{sec:convergence}

One of the key properties of an interactive segmentation algorithm is convergence to sufficient accuracy with an increasing number of clicks.  Previous works \cite{jang2019interactive, kontogianni2020continuous, sofiiuk2020f} improve the convergence using inference time optimization schemes, that force the predictions to match with input clicks. Our method is free from any refinement schemes and makes use of the proposed architecture and the iterative training scheme to achieve the desired level of convergence.

Sofiiuk~\etal~\cite{sofiiuk2020f} introduced the evaluation protocol for analysis of convergence for interactive segmentation methods. The main aspect of this protocol is the NoC\textsubscript{100} evaluation metric, which is similar to the NoC described earlier but has the clicks limit set to 100. Such size of interactive feedback should give an algorithm enough information to converge. In case the algorithm can not reach the IoU threshold, the image is supposedly too hard for the method to handle, therefore increasing the click limit is unlikely to change it.

Table~\ref{tab:convergence} shows the comparison of the the results of BRS \cite{jang2019interactive}, f-BRS-B \cite{sofiiuk2020f} and the proposed method with HRNet18+OCR backbone, reporting NoC\textsubscript{100}, number of images for which the 90\% threshold on IoU was not achieved after 20 and 100 clicks. All models are trained on the SBD \cite{hariharan2011semantic} dataset for a fair comparison.

% Table: convergence analysis
\begin{table}
\footnotesize
\begin{center}
\caption{Convergence analysis on Berkeley, SBD and DAVIS. We report the number of images that were not correctly segmented after 20 and 100 clicks and the NoC\textsubscript{100}@90 metric. All models are trained on SBD.}
\begin{tabular}{l|l|c|c|c}
\hline 
Dataset & Model & 
\begin{tabular}{@{}c@{}}\#images \\ $\geq$20 \end{tabular} & 
\begin{tabular}{@{}c@{}}\#images \\ $\geq$100 \end{tabular} & 
\begin{tabular}{@{}c@{}} NoC\textsubscript{100} \\  @90 \end{tabular} \\
\hline 
\hline
\multirow{4}{*}{Berkeley} & BRS \cite{jang2019interactive}  & 10 & 2 & 8.77 \\
                          & f-BRS-B \cite{sofiiuk2020f}     &  \textbf{2} & \textbf{0} & \textbf{4.47} \\
                          & HRNet-18                        &  7 & 3 & 7.10 \\
                          & HRNet-18 ITER-M                 &  3 & 2 & 4.89 \\
\hline
\multirow{4}{*}{DAVIS}    & BRS \cite{jang2019interactive}  & 77 & 51 & 20.98 \\
                          & f-BRS-B \cite{sofiiuk2020f}     & 78 & 50 & 20.70  \\
                          & HRNet-18                        & 67 & 50 & 19.83 \\
                          & HRNet-18 ITER-M                 & \textbf{57} & \textbf{44} & \textbf{18.42} \\
\hline
\multirow{3}{*}{SBD}      & f-BRS-B \cite{sofiiuk2020f}     & 1466 & 265 & 14.98 \\
                          & HRNet-18                        & 1051 & 450 & 13.62 \\
                          & HRNet-18 ITER-M                 &  \textbf{671} & \textbf{215} & \textbf{9.52} \\
\hline

\end{tabular}
\label{tab:convergence}
\end{center}
\end{table}

\subsection{Ablation Studies}
\label{sec:ablation}

% Table: iterative ablation
\begin{table}
\footnotesize
\begin{center}
\caption{Ablation studies on the maximum number of iterations $N_{iters}$ in the iterative sampling. All the results are reported for the model with the HRNet-18+OCR backbone trained on COCO+LVIS.}
\begin{tabular}{c|c|c|c|c}
\hline 
\multirow{2}{*}{$N_{iters}$} & Prev &  \multicolumn{3}{c}{NoC\textsubscript{20}@90} \\
\cline{3-5}
{} & {Mask} & Berkeley & DAVIS & SBD \\
\hline 
\hline

3 & -- & 2.38 & 5.92 & 6.49 \\
3 & + & 2.26 & 5.74 & 6.06 \\
\hline
1 & + & 2.57 & 5.81 & 6.15 \\
2 & + & 2.48 & \textbf{5.70} & 6.10 \\
3 & + & \textbf{2.26} & 5.74 & 6.06 \\
4 & + & 2.52 & 6.03 & \textbf{6.04} \\
5 & + & 2.49 & 5.98 & 6.24 \\
6 & + & 2.55 & 6.11 & 6.82 \\
\hline

\end{tabular}
\label{tab:iterative}
\end{center}
\end{table}

\textbf{Network architecture ablations.}
In Section \ref{sec:revising_network_and_brs}, we discuss different architectural choices for interactive segmentation networks. First, we explore different strategies of feeding encoded clicks to the model (DMF, Conv1E and Conv1S). We find that the HRNet-18 and ResNet-34 models with Conv1S show better performance in general and we use it in all further experiments. Then, we evaluate different clicks encoding strategies. We find that changing the distance transform encoding to the disk encoding significantly improves results of both HRNet-18 and ResNet-34, which confirms the findings of Benenson~\etal~\cite{benenson2019large} whose model performed best with disks with a radius of 3. However, our experiments show that a radius of 5 is better than 3. The results of all ablations can be found in Table~\ref{tab:ablation_architecture}.

In addition, these experiments clearly demonstrate the superiority of HRNet-18 over DeepLabV3+ with ResNet-34, especially when the former has several times fewer parameters and FLOPs than the latter (the number of parameters and FLOPs can be found in Table~\ref{tab:arch_comparison}). We use HRNet with the Conv1S input scheme and encode clicks with disks of radius 5 in all other experiments.

% Figure: a comparison between non-iterative and iterative models with and w/o prev mask
\begin{figure}[h]

\centering
\includegraphics[width=.48\textwidth]{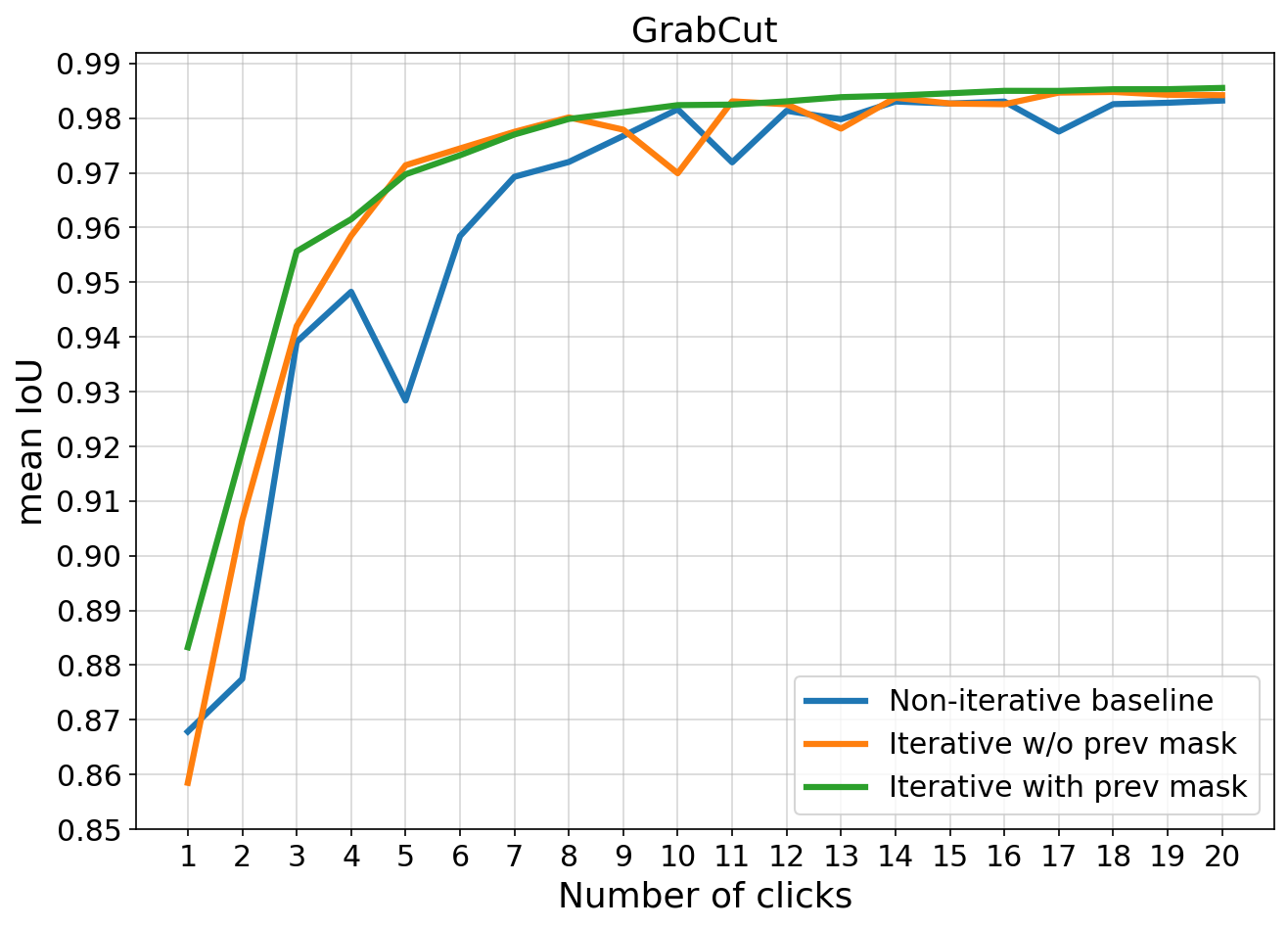}
\includegraphics[width=.48\textwidth]{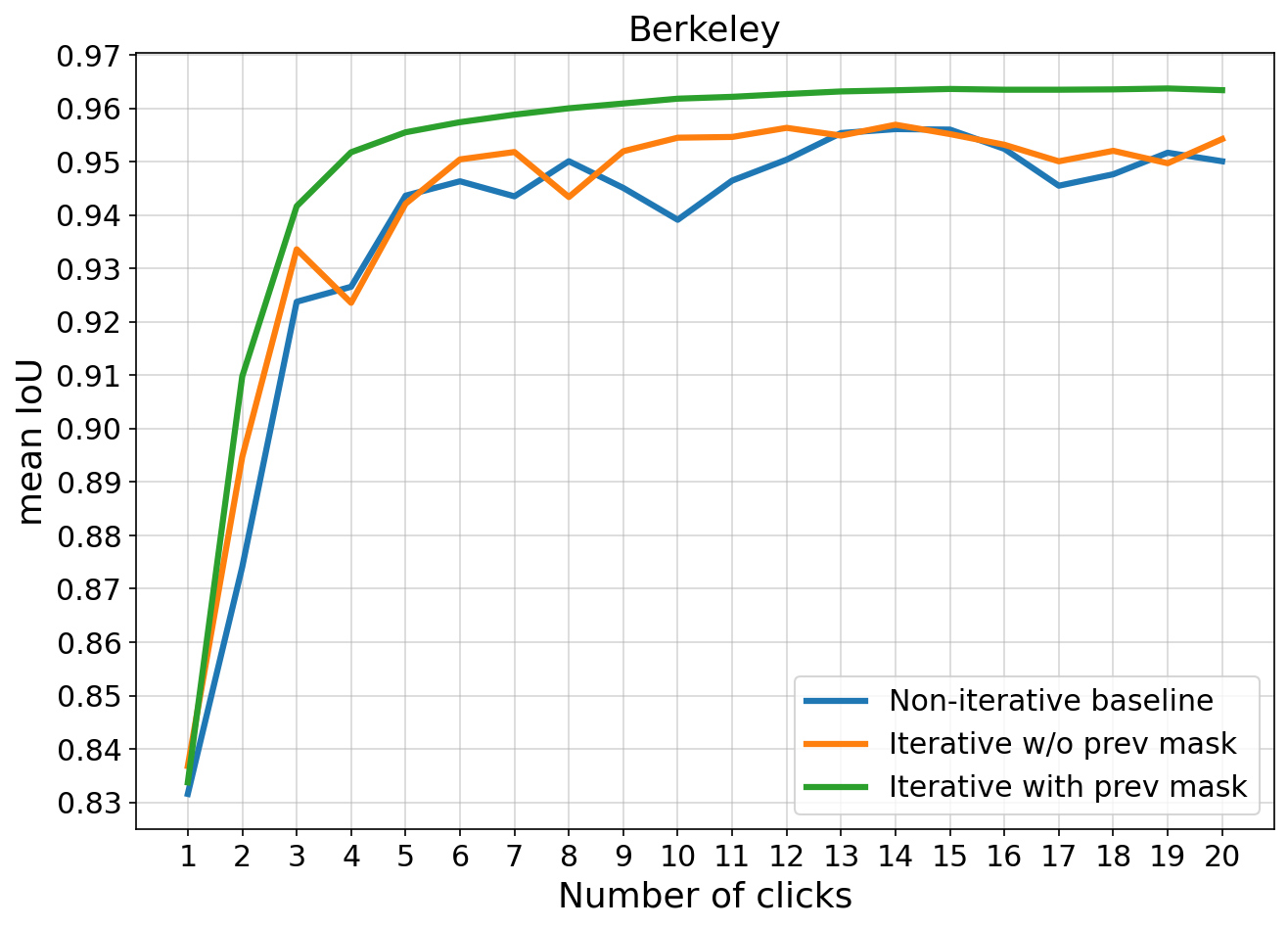}
\includegraphics[width=.48\textwidth]{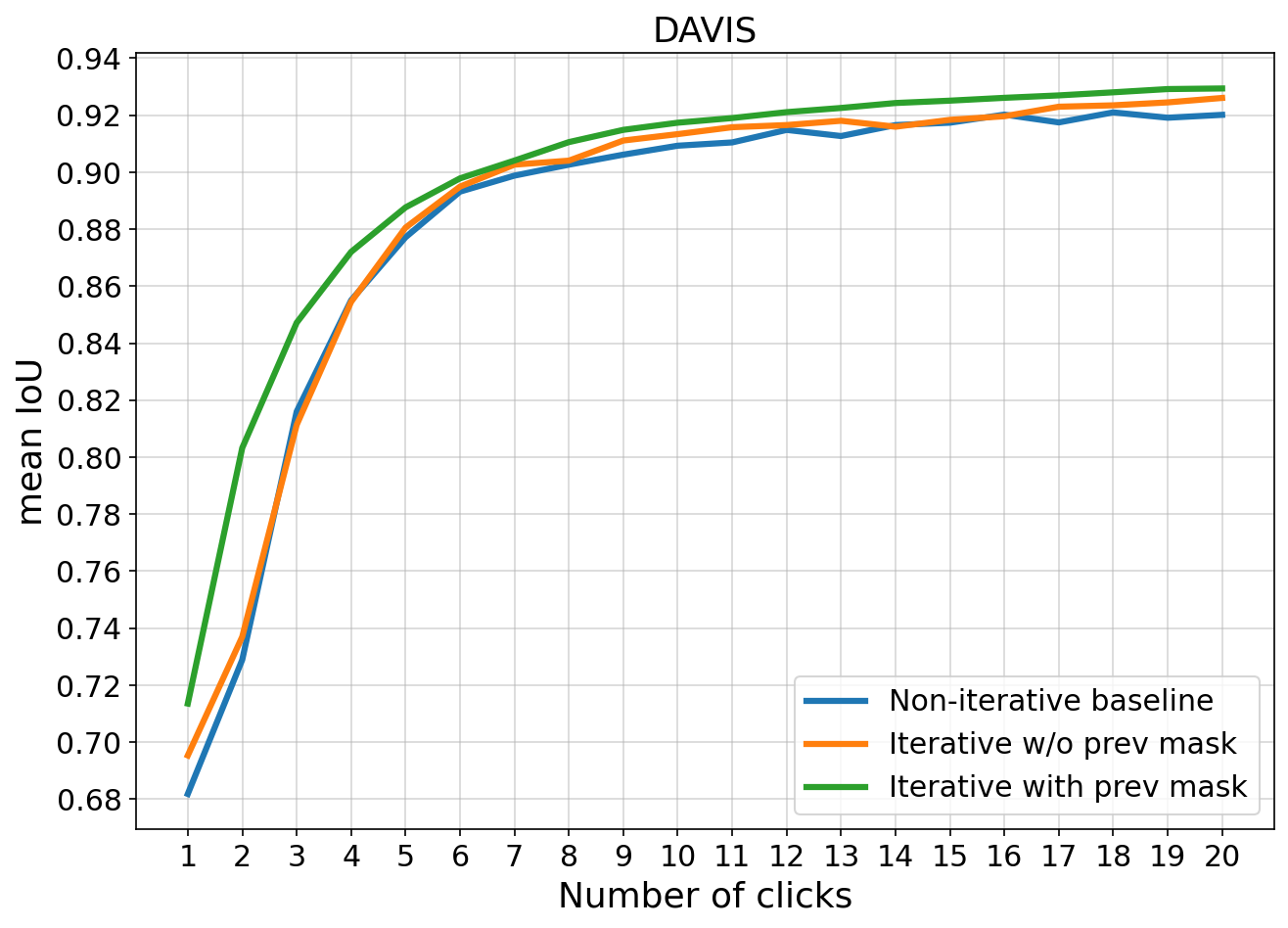}
\includegraphics[width=.48\textwidth]{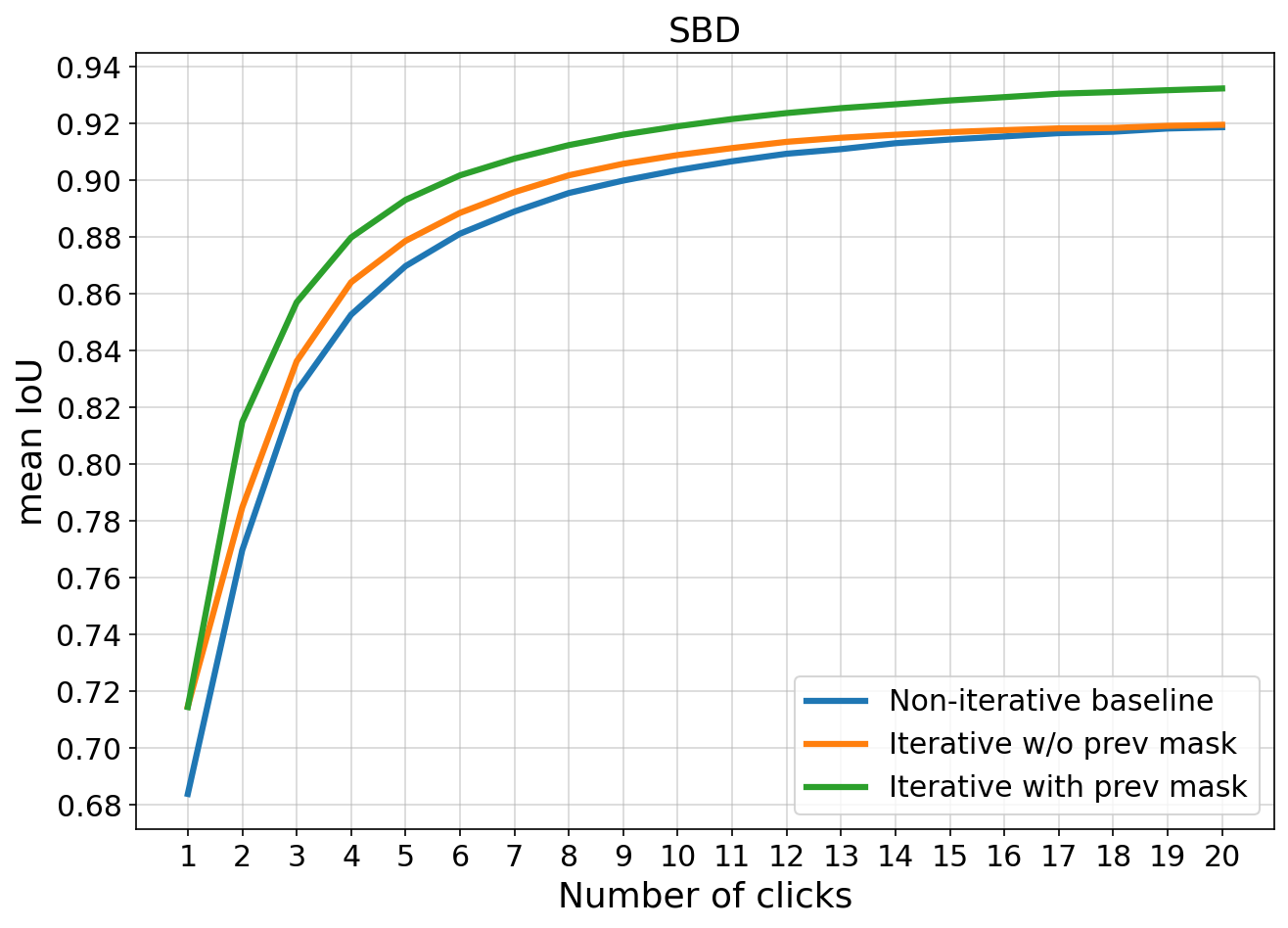}

\caption{Mean IoU@$k$ for varying number of clicks $k$ on GrabCut, Berkeley, DAVIS and SBD. The iterative model that takes a mask from a previous step is much more stable and converges to a better~IoU. All the results are reported for the model with the HRNet-18+OCR backbone trained on COCO+LVIS, iterative models are trained with $N_{iters}=3$. }
\label{fig:ious_ablation_mask}
\end{figure}

\textbf{Training datasets comparison.} To further research the findings from Section~\ref{sec:dataset_for_is}, we evaluate the models trained on each of the six common segmentation datasets: ADE20k \cite{zhou2017scene}, OpenImages \cite{benenson2019large, kuznetsova2018open}, SBD \cite{hariharan2011semantic}, Pascal~VOC \cite{everingham2010pascal} augmented with labels from SBD (denoted as Pascal~VOC+SBD), LVIS \cite{gupta2019lvis}, COCO \cite{lin2014microsoft} and also the COCO+LVIS dataset, proposed in Section~\ref{sec:coco_lvis}. We provide results for non-iterative baseline methods with HRNet-18+OCR and ResNet-34 backbones, trained with NFL loss. We report NoC@90 on Berkeley, SBD and DAVIS in Table~\ref{tab:ablation_datasets}. The performance on COCO and COCO+LVIS is relatively close, but the model trained on the proposed dataset generalizes better due to wider class distribution in the training set. Another observation is that the models trained on SBD and Pascal~VOC+SBD show the best performance on the SBD dataset in terms of NoC@90, which is most likely caused by the similar distribution of training and testing sets. Nonetheless, these models are inferior to the model trained on the COCO+LVIS dataset in terms of performance on other benchmarks.
All models trained for dataset comparison share the same training parameters and augmentations, that are described at the beginning of Section~\ref{sec:experiments}. We train the models for the following number of epochs on each dataset: 180 for ADE20k, 5 for OpenImages, 120 for SBD and Pascal~VOC+SBD, 55 for LVIS and COCO+LVIS, 40 for COCO. We denote one pass through each image of the dataset as an epoch.

\textbf{Loss functions comparison.} We compare four loss functions that were used for training segmentation methods in recent works: binary cross entropy (BCE) loss, focal loss (FL) \cite{lin2017focal}, soft IoU loss \cite{rahman2016optimizing} and normalized focal loss (NFL) \cite{sofiiuk2019adaptis}. We evaluate the performance of the baseline HRNet-18+OCR model trained on the COCO+LVIS dataset, described in Section~\ref{sec:coco_lvis}. The results are presented in Table~\ref{tab:ablation_loss}. We provide only NoC@90 on all standard datasets for simplicity. The evaluation results support the reasoning in Section~\ref{sec:normalized_focal_loss}, demonstrating that training with NFL leads to better accuracy and convergence on all 4 datasets.

\textbf{Iterative training ablations.}
The main hyperparameter of our iterative sampling scheme is the maximum number of iteratively sampled clicks $N_{iters}$ in addition to some set of randomly sampled clicks. Our experiments show that $N_{iters}=3$ is an optimal value of that parameter. Surprisingly, too high values (> 4) lead to instability during training and to worse results. We had to train models with $N_{iters}=5,6$ for several times, as they collapsed after 10-20 epochs of training on COCO+LVIS and showed poor results. We provide the ablation study in Table~\ref{tab:iterative}. 

We study the impact of feeding a mask from the previous click to the iterative model with $N_{iters}=3$. Apart from metrics improvement, we observe substantial improvement of the stability of the model when new clicks are added. As shown in Figure~\ref{fig:ious_ablation_mask}, both the iterative model without a previous mask and the non-iterative model sometimes have drops of mean IoU when the number of clicks increases. It indicates that adding a new click during the process of interactive segmentation can even make the mask worse, contradicting user expectations. This effect is better illustrated on mean IoU plots for small datasets such as GrabCut or Berkeley, as the smoothness of the curve is proportional to the dataset size. When the model relies on a mask from a previous iteration, it can take into account the segmentation result of a previous iteration and avoid unexpected collapse of the current mask. Moreover, it allows to modify existing masks without additional effort. We discover that the trained model can be successfully initialized with an external inaccurate mask without the history of previous clicks to correct the errors of this mask.  Several examples of applying our to this use case are shown in Figure~\ref{fig:external_mask_correction}.

% Figure: Visualisation on Berkeley
\newpage
\begin{figure}[!ht]
    \centering
    \includegraphics[width=0.98\linewidth]{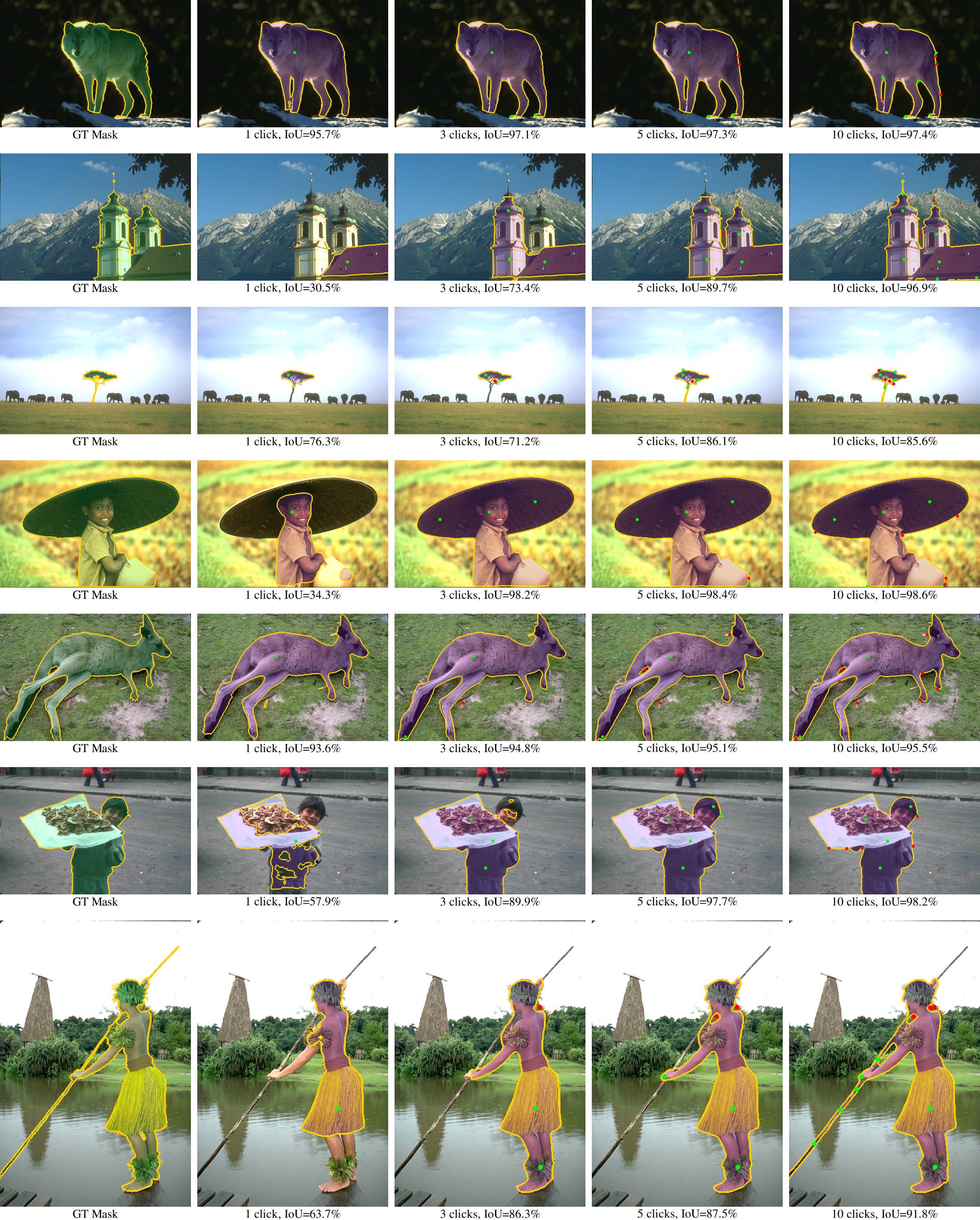}
    \caption{Visualization of interactive segmentation for the Berkeley images with a different number of clicks fed to the HRNet-18 ITER-M model and obtained by the NoC evaluation procedure \cite{xu2016deep}. Green and red dots denote positive and negative clicks, respectively. There are only 2 images from Berkeley on which our model does not converge to 90\% IoU in 20 clicks. One of them is shown in the third row.}
    \label{fig:is_vis_berkeley}
\end{figure}

\subsection{Comparison with Previous Works}
The quantitative results are summarized in Table~\ref{tab:evaluation_all}. We notice that even the proposed baseline model with HRNet18+OCR backbone outperforms all previous methods. The proposed iterative method that uses a mask from the previous click sets a new state-of-the-art in interactive segmentation on all five benchmarks. Another notable fact is that the smallest among the proposed models with HRNet18s+OCR backbone performs on par with heavier ones, making it possible to use the proposed method on devices with low computational capability.

\begin{figure}[ht]
    \centering
    \includegraphics[width=.48\textwidth]{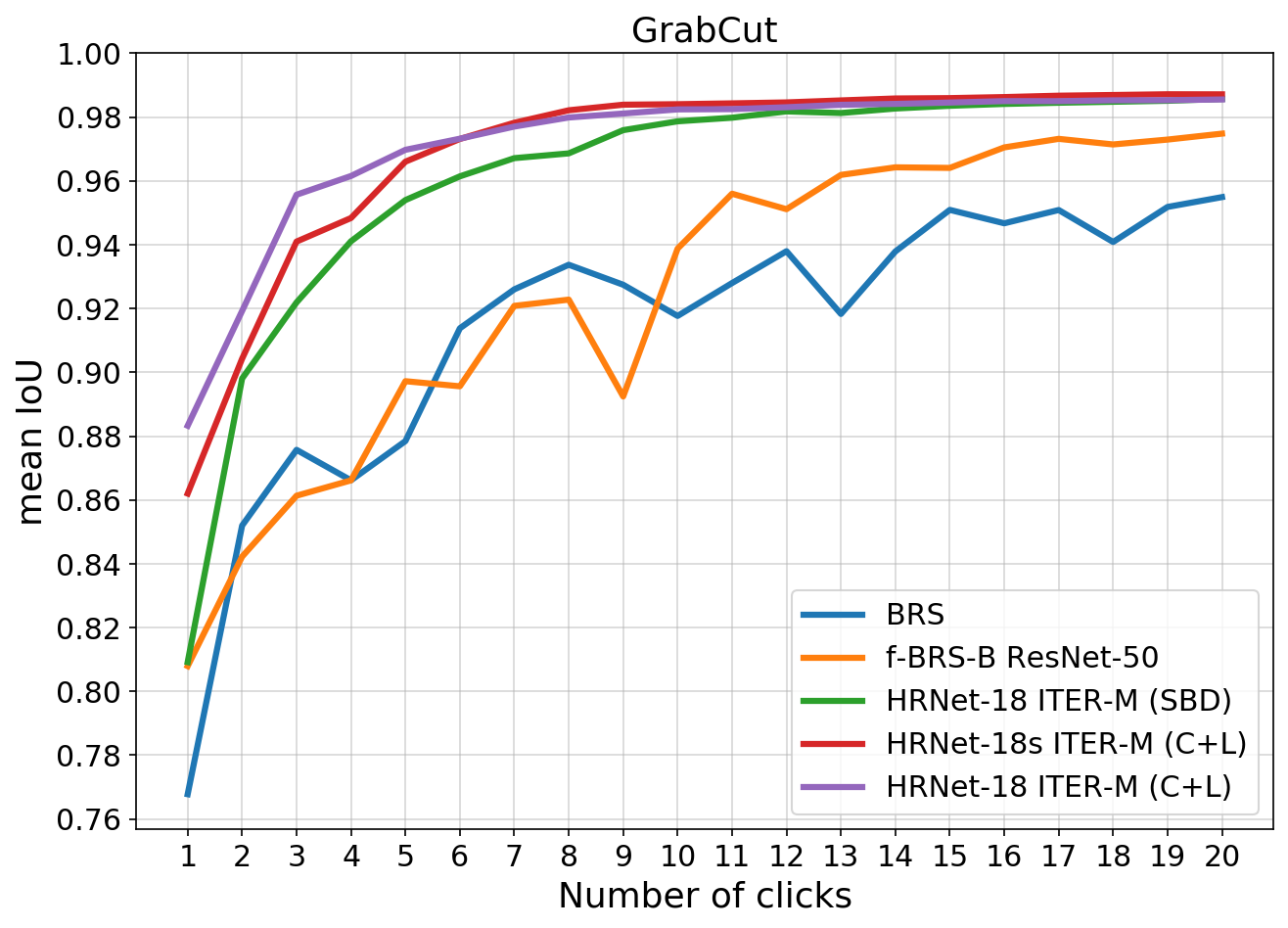}
    \includegraphics[width=.48\textwidth]{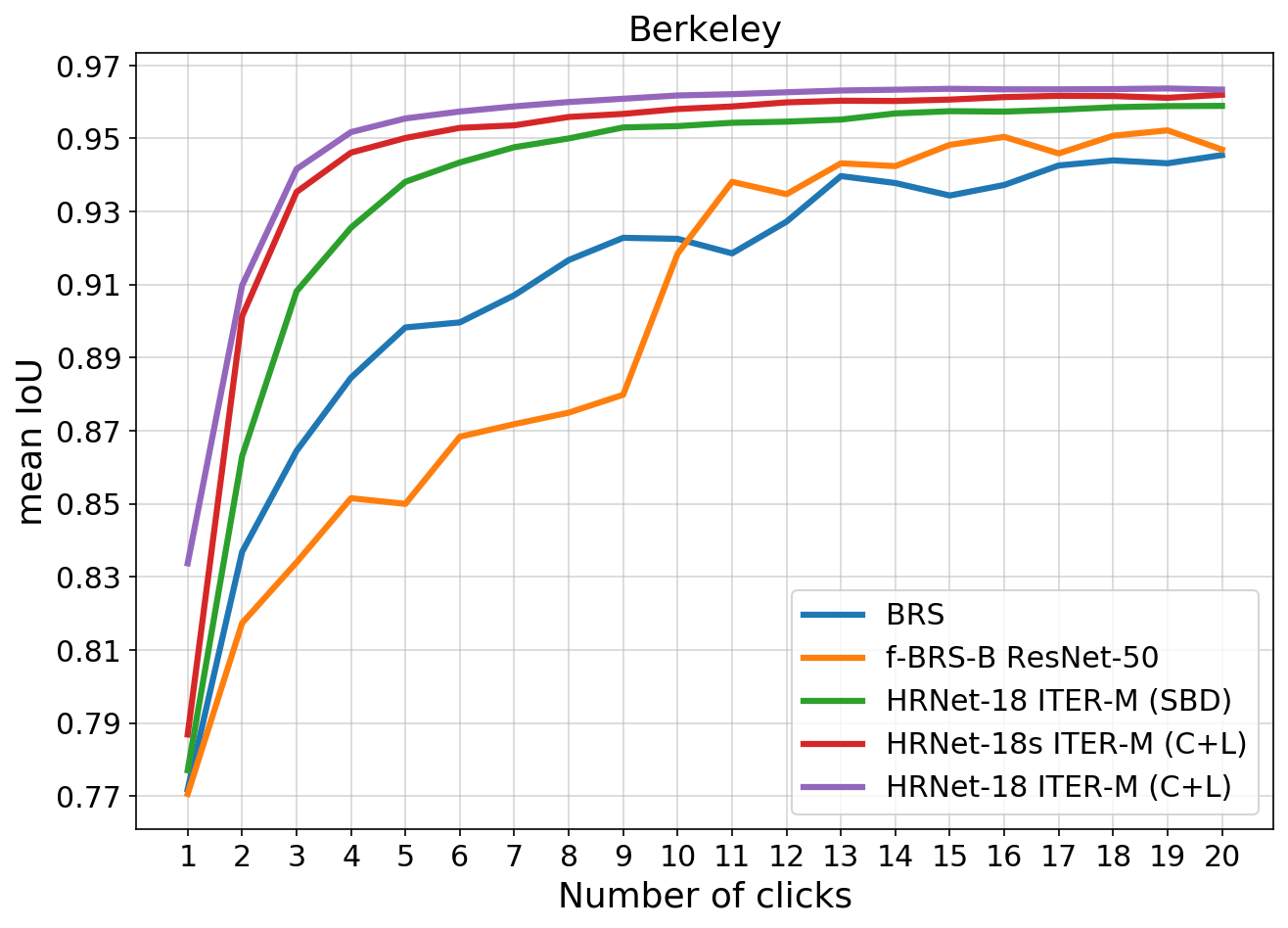}
    \includegraphics[width=.48\textwidth]{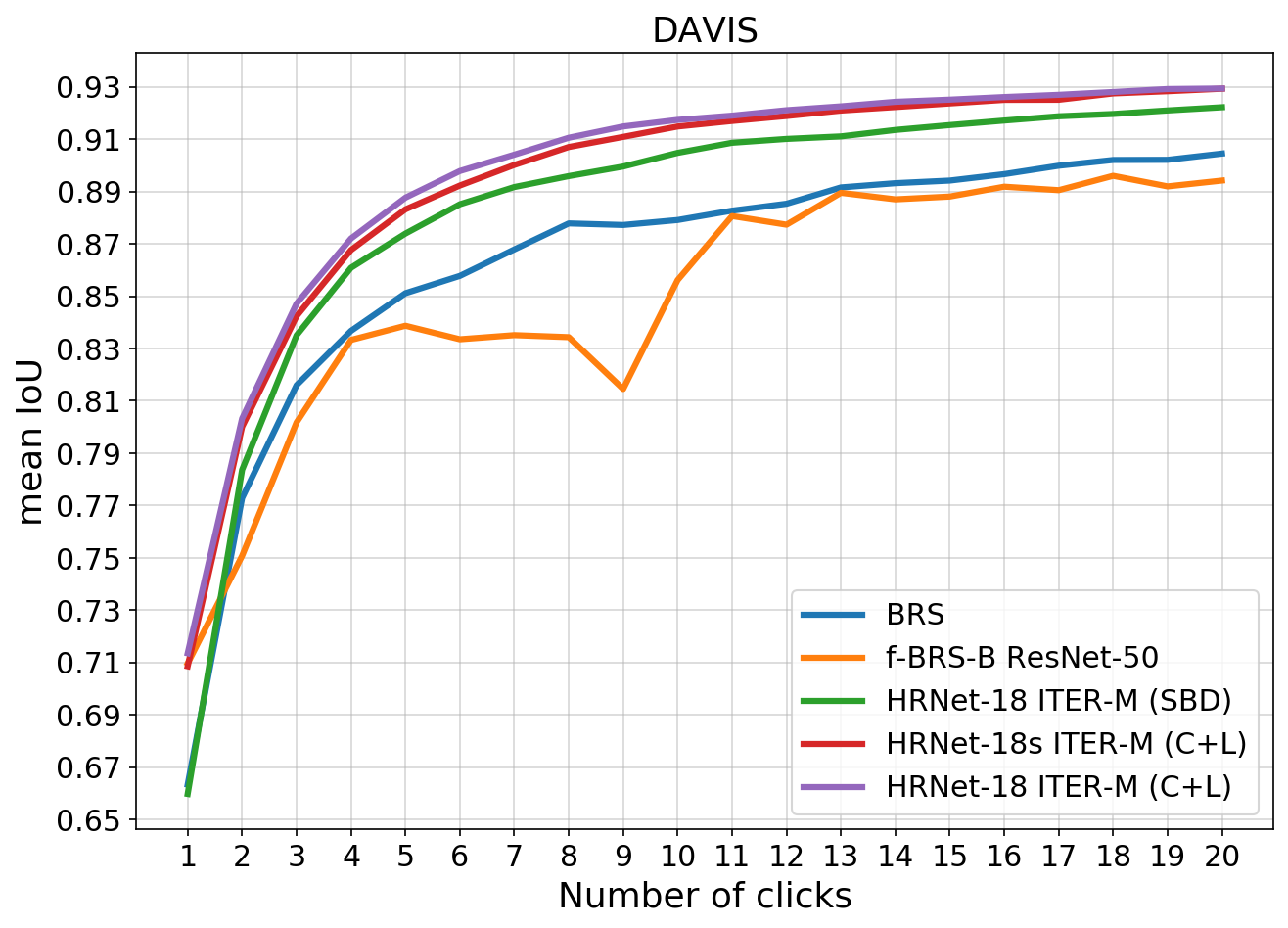}
    \includegraphics[width=.48\textwidth]{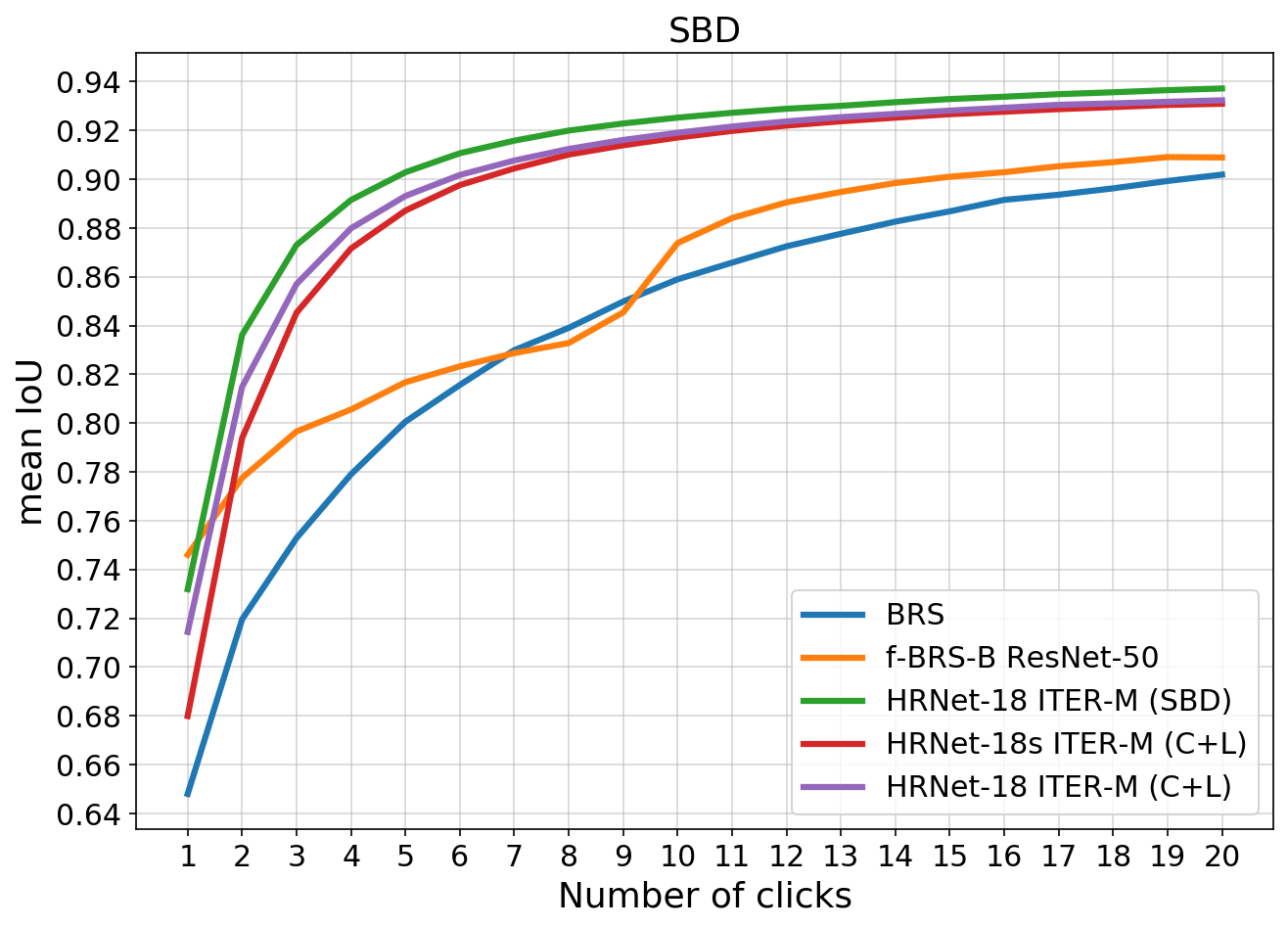}
    
    \caption{Mean IoU@$k$ for varying number of clicks $k$ on GrabCut, Berkeley, DAVIS and SBD. The iterative models (denoted as ITER-M) show stable performance without accuracy drops and converge to a better IoU. Names of the training datasets are enclosed in parentheses. "C+L" stands for COCO+LVIS.}
    \label{fig:miou_plots}
\end{figure}

We provide plots of mean IoU with respect to the number of clicks for the GrabCut, Berkeley, DAVIS and SBD datasets in Figure~\ref{fig:miou_plots}. The proposed method outperforms previous state-of-the-art works and shows great improvement in accuracy and stability while avoiding accuracy drops after any click and converging to a better result.

We also visualise evaluation process of the proposed iteratively trained HRNet-18 model on some images from the Berkeley dataset in Figure~\ref{fig:is_vis_berkeley}. 

% Table: comparison with previous works
\begin{table}
\small
\begin{center}
\caption{Evaluation results on GrabCut, Berkeley, SBD, DAVIS and Pascal VOC datasets. The best and the second best results are set in bold and underlined, respectively. "H18s", "H18" and "H32" stand for "HRNet-18s", "HRNet-18" and "HRNet-32", respectively. Our models with the "IT-M" suffix are iterative models that take a mask from a previous step with $N_{iters}=3$, otherwise they are our non-iterative baselines. The name of the training dataset is indicated below the word "Ours". "C+L" stands for COCO+LVIS.}
\begin{tabular}{ll|c|c|c|c|c|c|c|c}
\hline
\multicolumn{2}{l|}{\multirow{2}{*}{Method}} & \multicolumn{2}{c|}{GrabCut} & Berkeley & \multicolumn{2}{c|}{SBD} & \multicolumn{2}{c|}{DAVIS} & {Pascal VOC} \\
\cline{3-10}
\multicolumn{2}{l|}{} & NoC@85 & NoC@90 & NoC@90 & NoC@85 & NoC@90 & NoC@85 & NoC@90 & NoC@85 \\
\hline
\hline
\multicolumn{2}{l|}{GC \cite{boykov2001interactive}}             &  7.98 & 10.00 & 14.22 & 13.60 & 15.96 & 15.13 & 17.41 & -- \\
\multicolumn{2}{l|}{GM \cite{gulshan2010geodesic}}               & 13.32 & 14.57 & 15.96 & 15.36 & 17.60 & 18.59 & 19.50 & -- \\
\multicolumn{2}{l|}{RW \cite{grady2006random}}                   & 11.36 & 13.77 & 14.02 & 12.22 & 15.04 & 16.71 & 18.31 & -- \\
\multicolumn{2}{l|}{ESC \cite{gulshan2010geodesic}}              &  7.24 &  9.20 & 12.11 & 12.21 & 14.86 & 15.41 & 17.70 & -- \\
\multicolumn{2}{l|}{GSC \cite{gulshan2010geodesic}}              &  7.10 &  9.12 & 12.57 & 12.69 & 15.31 & 15.35 & 17.52 & -- \\
\hline
\multicolumn{2}{l|}{DIOS with GC \cite{xu2016deep}}              &   --  &  6.04 &  8.65 &   --  &   --  &   --  &   --  & 6.88 \\
\multicolumn{2}{l|}{Latent diversity \cite{li2018interactive}}   &  3.20 &  4.79 &   --  &  7.41 & 10.78 &  5.05 &  9.57 & -- \\
\multicolumn{2}{l|}{RIS-Net \cite{liew2017regional}}             &   --  &  5.00 &  6.03 &   --  &   --  &   --  &   --  & 5.12 \\
\multicolumn{2}{l|}{ITIS \cite{mahadevan2018iteratively}}        &   --  &  5.60 &   --  &   --  &   --  &   --  &   --  & 3.80 \\
\multicolumn{2}{l|}{CAG \cite{majumder2019content}}              &   --  &  3.58 &  5.60 &   --  &   --  &   --  &   --  & 3.62 \\
\multicolumn{2}{l|}{BRS \cite{jang2019interactive}}              &  2.60 &  3.60 &  5.08 &  6.59 &  9.78 &  5.58 &  8.24 & -- \\
\multicolumn{2}{l|}{FCA-Net (SIS) \cite{lin2020interactive}}     &   --  &  2.08 &  3.92 &   --  &   --  &   --  &  7.57 & 2.69 \\
\multicolumn{2}{l|}{IA+SA \cite{kontogianni2020continuous}}      &   --  &  3.07 &  4.94 &   --  &   --  &  5.16 &   --  & 3.18 \\
\multicolumn{2}{l|}{f-BRS-B \cite{sofiiuk2020f}}                 &  2.50 &  2.98 &  4.34 &  5.06 &  8.08 &  5.39 &  7.81 & -- \\
\hline
\hline
\multirow{2}{*}{\shortstack{Ours\\ \footnotesize{SBD}}} 
                      & H18                                      & 1.96 & 2.41 & 3.95 & 4.12 & 6.66 & 5.08 & 7.17 & 2.94 \\
                      & H18 IT-M                                 & 1.76 & 2.04 & 3.22 & \textbf{3.39} & \textbf{5.43} & 4.94 & 6.71 & \underline{2.51} \\
\hline
\multirow{4}{*}{\shortstack{Ours\\ \footnotesize{C+L}}}
                      & H18                                      & 1.54 & 1.70 & 2.48 & 4.26 & 6.86 & 4.79 & 6.00 & 2.59 \\
                      & H18s IT-M                                & 1.54 & 1.68 & 2.60 & 4.04 & 6.48 & 4.70 & 5.98 & 2.57 \\
                      & H18 IT-M                                 & \textbf{1.42} & \textbf{1.54} & \underline{2.26} & 3.80 & 6.06 & \underline{4.36} & \underline{5.74} & \textbf{2.28} \\
                      & H32 IT-M                                 & \underline{1.46} & \underline{1.56} & \textbf{2.10} & \underline{3.59} & \underline{5.71} & \textbf{4.11} & \textbf{5.34} & 2.57 \\
\hline

\end{tabular}
\label{tab:evaluation_all}
\end{center}
\end{table}

% Figure: mIoU plots

\section{Conclusion}

Most of recent studies on interactive segmentation propose models that heavily rely on additional inference-time optimization schemes. In our work, we have demonstrated that a pure feedforward model with a modern backbone architecture can achieve or even surpass current state-of-the-art results. We have introduced a new model able to modify existing segmentation masks as well as segment new objects without any prior masks. It sets a new state-of-the-art on all common interactive segmentation benchmarks.

Our experiments have proved that a training dataset has a major impact on the model performance. We have proposed to combine the two existing instance segmentation datasets, COCO~\cite{lin2014microsoft} and LVIS~\cite{gupta2019lvis}, and use it for interactive segmentation. The resulting large and diverse dataset with high-quality annotations has significantly improved the generalization ability of our model. Training on this dataset allows to push the state-of-the-art results achieved by our model even further.

\noindent\textbf{Acknowledgment.} We thank Julia Churkina for her assistance with editing and for comments that greatly improved the manuscript.

\bibliographystyle{model1-num-names}

\bibliography{cas-refs}


% Author biography without author photo.
% Author biography. Author biography. Author biography.
% Author biography. Author biography. Author biography.
% Author biography. Author biography. Author biography.
% Author biography. Author biography. Author biography.
% Author biography. Author biography. Author biography.
% Author biography. Author biography. Author biography.
% Author biography. Author biography. Author biography.
% Author biography. Author biography. Author biography.
% Author biography. Author biography. Author biography.

\end{document}